\documentclass{article}

\usepackage[preprint]{neurips_2025}

\usepackage[utf8]{inputenc} 
\usepackage[T1]{fontenc}    
\usepackage{hyperref}       
\usepackage{url}            
\usepackage{booktabs}       
\usepackage{amsfonts}       
\usepackage{nicefrac}       
\usepackage{microtype}      
\usepackage{xcolor}         
\usepackage{amsmath}
\usepackage{cleveref}
\usepackage{graphicx}
\usepackage{pdfpages}

\title{A Semantic Parsing Framework for End-to-End Time Normalization}

\author{%
  Xin Su \\
  Intel Labs \\
  \texttt{xin.su@intel.com} \\
  \And
  Sungduk Yu \\
  Intel Labs \\
  \texttt{sungduk.yu@intel.com} \\
  \And
  Phillip Howard \\
  Thoughtworks \\
  \texttt{phillip.howard@thoughtworks.com} \\
  \And
  Steven Bethard \\
  University of Arizona \\
  \texttt{bethard@arizona.edu} \\
}

\begin{document}

\maketitle

\begin{abstract}
Time normalization is the task of converting natural language temporal expressions into machine-readable representations. It underpins many downstream applications in information retrieval, question answering, and clinical decision-making. Traditional systems based on the ISO-TimeML schema limit expressivity and struggle with complex constructs such as compositional, event-relative, and multi-span time expressions. In this work, we introduce a novel formulation of time normalization as a code generation task grounded in the SCATE framework, which defines temporal semantics through symbolic and compositional operators. We implement a fully executable SCATE Python library and demonstrate that large language models (LLMs) can generate executable SCATE code. Leveraging this capability, we develop an automatic data augmentation pipeline using LLMs to synthesize large-scale annotated data with code-level validation. Our experiments show that small, locally deployable models trained on this augmented data can achieve strong performance, outperforming even their LLM parents and enabling practical, accurate, and interpretable time normalization.
\end{abstract}

\section{Introduction}
Time normalization refers to the task of converting temporal expressions in natural language into machine-readable formats. For example, the phrase \textit{``three days ago''} spoken on August 25, 2024, should be normalized to \texttt{2024-08-22}. Time normalization plays a crucial role in a variety of temporal reasoning applications, including literature study \citep{fischer2015does}, question answering \citep{su-etal-2023-fusing}, event analysis \citep{vossen2016newsreader}, and clinical decision-making \citep{lin2015automatic}.

Most existing time normalization systems \citep{kim-etal-2020-time, shwartz-2022-good, lange-etal-2023-multilingual} are based on the ISO-TimeML framework \citep{pustejovsky-etal-2010-iso}. These systems typically follow a two-stage pipeline: first identifying temporal expressions in text, then classifying them into predefined normalized representations (e.g., mapping ``noon'' to a specific time of day). While effective for standard expressions, this approach is fundamentally limited by the rigidity and restricted expressivity of the ISO-TimeML schema. As pointed out by \citet{bethard-parker-2016-semantically}, ISO-TimeML-based systems cannot effectively handle several classes of temporal expressions:

\begin{enumerate}
    \item \textbf{Multi-span expressions} that span across calendar units, such as \textit{``every Monday for the past three weeks''};
    \item \textbf{Event-relative expressions}, such as \textit{``three weeks postoperative''}, where "three weeks" is relative to "surgery";
    \item \textbf{Compositional expressions} that involve multiple temporal constructs, as in \textit{``May 22, 1995 ... and the following month''}, where the latter phrase is semantically dependent on the former.
\end{enumerate}

\begin{figure*}
    \centering
    \includegraphics[scale=0.2]{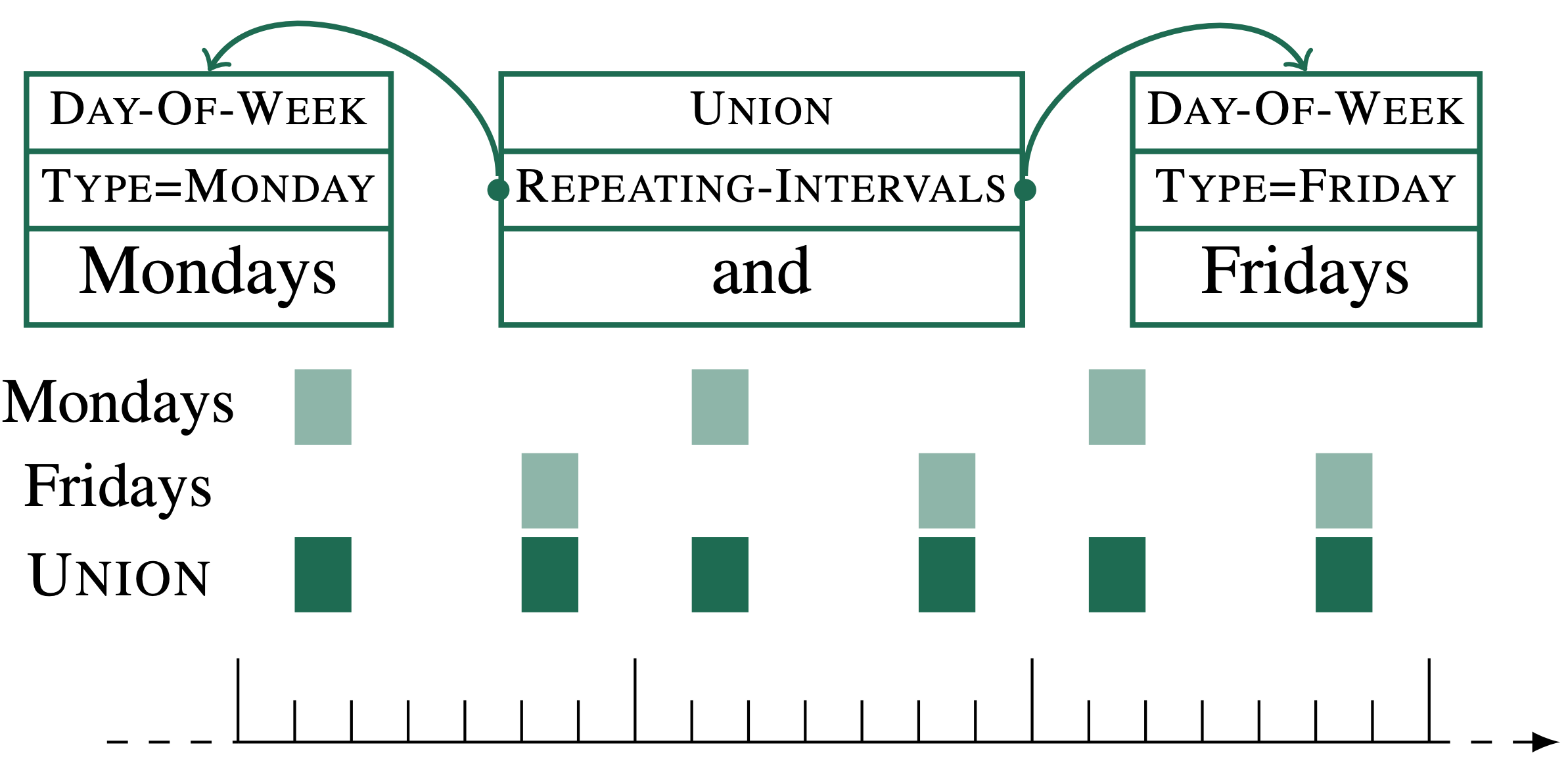}
    \caption{UNION annotation: Mondays and Fridays}
    \label{fig:scate-example}
\end{figure*}

To overcome these limitations, \citet{bethard-parker-2016-semantically} propose the Semantically Compositional Annotation of Temporal Expressions (SCATE) framework. SCATE represents temporal expressions through compositional and symbolic semantics using a rich set of temporal operators (e.g., \texttt{Union}, \texttt{Intersection}, \texttt{RepeatingInterval}). For instance, the expression \textit{``Mondays and Fridays''} is represented as:

\begin{align*}
\texttt{Union(} 
    &\texttt{RepeatingInterval(DayOfWeek(Type=Monday)),} \\
    &\texttt{RepeatingInterval(DayOfWeek(Type=Friday))} 
\texttt{)}
\end{align*}

As illustrated in \Cref{fig:scate-example}, this operation produces all occurrences of Mondays and Fridays along the timeline. In this representation, both \textit{Mondays} and \textit{Fridays} are modeled as \texttt{RepeatingInterval} objects, and the conjunction \textit{and} is captured using the \texttt{Union} operator.  This compositional design allows SCATE to handle a broader range of temporal expressions and produce precisely defined, timeline-anchored intervals. Despite its expressivity, existing SCATE-based systems \citep{laparra-etal-2018-characters,xu-etal-2019-pre} rely on complex, multi-stage pipelines. These systems treat SCATE as an annotation schema, training separate models to identify its atomic concepts and then applying handcrafted rule systems to reconstruct full interpretations. This approach results in high runtime costs, reduced maintainability, and limited deployability in real-world applications.

In this work, we propose a novel, end-to-end formulation of time normalization as a code generation task. Our key insight is to implement the full SCATE framework as an executable Python library, where all core temporal concepts and operations are mapped to Python classes and functions. This enables us to directly generate executable SCATE code from natural language inputs and deterministically compute their normalized time intervals.

Our main contributions are as follows:

\begin{itemize}
    \item We design and implement a complete Python library that faithfully captures all concepts in SCATE, making the semantics of each time expression interpretable and executable.
    
    \item We construct detailed prompting strategies to leverage large language models (LLMs) for data augmentation, generating 10$\times$ more labeled examples than existing annotated datasets, with automatic validation through code execution.
    
    \item We demonstrate that small, locally deployable models ($\leq$1B parameters) can be trained on the augmented dataset to generate SCATE code with competitive performance, enabling practical deployment of expressive time normalization systems.
\end{itemize}

\section{Related Works}

\subsection{Time Normalization}
The vast majority of existing time normalization methods are built upon the ISO-TimeML framework. Early systems, such as HeidelTime \citep{strotgen-gertz-2010-heideltime} and SUTime \citep{chang-manning-2012-sutime}, adopt rule-based approaches to convert recognized temporal expressions into the standardized format defined by ISO-TimeML. More recently, transformer-based models have been introduced for this task. For instance, \citet{shwartz-2022-good}, \citet{lange-etal-2023-multilingual}, and \citet{kim-etal-2020-time} train transformer-based classification models to map identified temporal expressions to predefined normalized time categories. In contrast, \citet{laparra-etal-2018-characters} and \citet{xu-etal-2019-pre} present the only complete time normalization pipeline grounded in the SCATE framework. Their approach involves using neural models, such as LSTMs, for temporal expression recognition, followed by a rule-based component to link these expressions to their final normalized forms. Distinct from all prior work, our method introduces a simple and practical end-to-end solution for time normalization formulated as a code generation task. By generating executable Python code, our approach directly maps textual temporal expressions to their normalized representations, offering a seamless and interpretable mechanism.

\subsection{Information Extraction via Code Generation}
Another related line of work gaining much recent attention is the representation of information extraction (IE) tasks using programming code or code-like structures. Instead of producing free-form text or sequences of labels, these approaches explicitly represent extracted information as structured code. This paradigm is particularly appealing in the era of powerful pretrained LLMs, which can effectively translate implicit semantic structures into explicit code-like or structured formats.

Recent studies demonstrated that code-based prompting allows efficient capturing of structured information. Code4Struct \citep{wang2022code4struct} and CodeIE \citep{li2023codeie} framed IE tasks explicitly as code generation problems, showing that code-specialized LLMs outperform natural-language LLMs in few-shot settings for IE tasks such as named entity recognition (NER) and relation extraction (RE). Expanding this further, Code4UIE \citep{guo2024retrieval} introduced retrieval-augmented code prompting, retrieving relevant few-shot examples, along with their universal IE approach.

An alternative inference-time strategy by \citet{geng2023grammar} adapted grammar-constrained decoding, explicitly enforcing output schema via formal grammars, thus achieving competitive IE performance without any model fine-tuning.

Complementing inference-time methods, another approach focuses on training specialized IE models. Doc2Dict \citep{townsend2021doc2dict} directly trained a generative T5 model to produce structured JSON outputs from documents, eliminating intermediate annotation steps unlike traditional pipeline-based models. More recently, KnowCoder \citep{li2024knowcoder} proposed a dedicated IE LLM, undergoing a two-phase training (schema-based code pretraining and schema-guided instruction tuning), significantly outperforming general-purpose LLMs. In subsequent work, KnowCoder-X \citep{zuo2025knowcoderxboostingmultilingualinformation} further extended this framework to multilingual IE tasks through cross-lingual alignment training, achieving state-of-the-art multilingual IE performance.

\begin{figure*}
    \centering
    \includegraphics[scale=0.5]{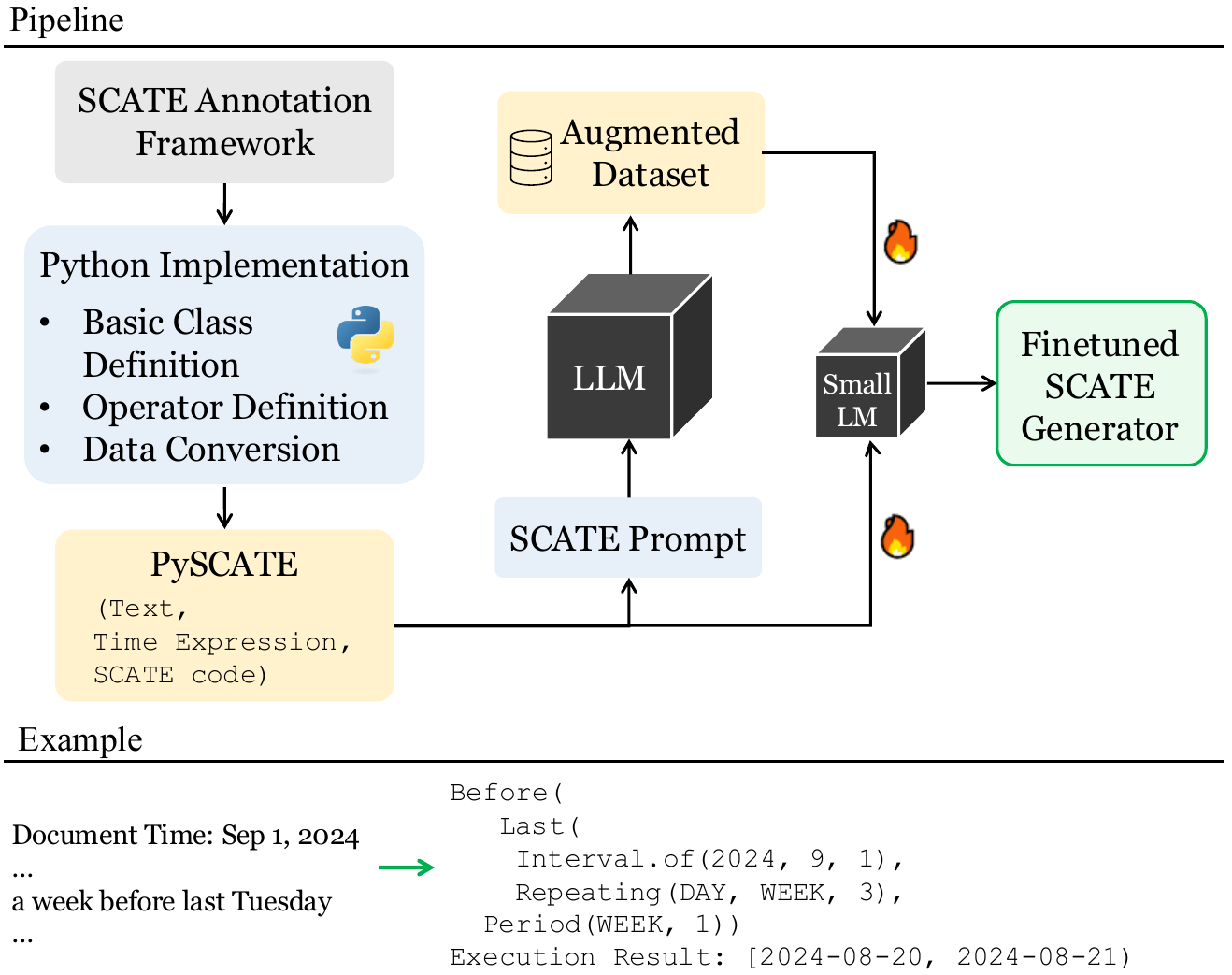}
    \caption{Overview of our approach }
    \label{fig:method-overview}
\end{figure*}

\section{Methodology}
\paragraph{Overview}
In this paper, we focus on the task of identifying temporal expressions from a given text and parsing them into the corresponding SCATE framework code. By deterministically executing the parsed structured code, we obtain time intervals anchored to a timeline. To accomplish this task, we first implement the concepts and operations defined in SCATE as fully executable Python objects packaged as a Python library, converting existing SCATE annotations into corresponding code representations. We then leverage our Python package to construct language model prompts for large-scale annotation of unlabeled text, generating additional annotations with quality-enhancing filtering mechanisms. Finally, we train small-scale language models on both the converted data and the additional LLM-annotated data for end-to-end time normalization code generation. We present an overview of our method in \Cref{fig:method-overview}.

\subsection{Task Definition}
Our proposed method is based on the SCATE temporal normalization framework. The fundamental principle of the SCATE framework is to represent complex temporal expressions compositionally, addressing the limitations of common temporal expression frameworks such as TimeML, which has limited expressivity. For instance, TimeML cannot represent expressions that cannot be aligned to a single calendar unit, such as "the past three summers."

The SCATE framework defines five key temporal concepts:
\begin{itemize}
    \item \textbf{Timeline}: An infinite sequence of time points to which temporal expressions (or events) can be anchored. Each time point is assumed to have second-level precision, e.g., 2015-08-03 09:35:47 represents a point on the timeline.
    
    \item \textbf{Interval}: A segment that can be precisely anchored to the timeline, defined by a starting point (inclusive) and an ending point (exclusive). For example, ``1990'' corresponds to $[1990$-$01$-$01\ 00$:$00$:$00, 1991$-$01$-$01\ 00$:$00$:$00)$.
    
    \item \textbf{Repeating interval}: A sequence of intervals on the timeline. For example, ``Friday'' refers to every Friday each week, representing a sequence of intervals that repeats infinitely.
    
    \item \textbf{Period}: An amount of time expressed as counts of standard time units. Periods are independent of the timeline. For example, ``10 weeks'' does not have specific start or end points.
    
    \item \textbf{Temporal operator}: A higher-order function that operates over periods, intervals, and repeating intervals to produce new temporal expressions. For example, the expression ``Saturdays in March'' involves two repeating intervals--``Saturdays'' and ``March''--combined via an \textsc{Intersection} operator. The result is a set of Saturday intervals occurring within the month of March, which can then be anchored to the timeline. For formal operator definitions, we refer the reader to the original SCATE paper \citep{bethard-parker-2016-semantically}.
\end{itemize}

In our work, we implement all SCATE temporal constructs as composable and executable Python objects. Formally, given a text $T$ containing $n$ time expressions $\{\text{timex}_1, \dots, \text{timex}_n\}$, our goal is to train a parameterized model $M$ that maps $T$ to a corresponding set of SCATE code representations $\{\text{code}_1, \dots, \text{code}_n\}$ such that:

\[
\text{Execute}(\text{code}_i) \rightarrow \text{Interval}_i \quad \text{for } i = 1, \dots, n
\]

where each $\text{Interval}_i$ is a normalized, timeline-anchored temporal interval.

This formulation enables the task to be approached as a structured code generation problem, with model outputs grounded in a formally defined temporal logic system.

\subsection{SCATE Code Representation}

\subsubsection{Base Class Definitions}

To faithfully represent the SCATE framework's temporal concepts, we implement a comprehensive object-oriented Python library. Our implementation centers around several base classes that directly correspond to SCATE's fundamental concepts. Our design adheres to two core principles: compositionality and executability. The compositional nature allows classes to be flexibly combined to represent complex temporal expressions, mirroring how most natural language constructs temporal references. Meanwhile, the executability principle ensures each object can be deterministically executed to produce concrete intervals on the timeline, with all classes implementing necessary addition and subtraction methods for direct interaction with Python \texttt{datetime} objects.

\paragraph{Interval Class.} The Interval class serves as the cornerstone of our implementation, directly embodying SCATE's interval concept. It represents a specific time span on the timeline with well-defined start and end points. Intervals can be created through various constructors, such as \texttt{Interval.of(1990)} to represent the year 1990, \texttt{Interval.of(1990, 1, 1)} for more specific dates like January 1, 1990, or using standard ISO format through \texttt{Interval.fromisoformat(``1990-01-01T00:00:00 1994-01-01T00:00:00'')}.

\paragraph{Unit Class.} We implement a Python enumeration to represent standard time units (e.g., \texttt{SECOND}, \texttt{MINUTE}, \texttt{HOUR}, \texttt{DAY}, \texttt{WEEK}, \texttt{MONTH}, \texttt{YEAR}, \texttt{CENTURY}). These units serve as building blocks for periods and provide utilities for truncating dates and calculating relative deltas.

\paragraph{Shift Class.} We introduce the Shift class as an abstract base class that captures the concept of movement along the timeline. The \texttt{Shift} class defines the interface for objects that can be added to or subtracted from time points to yield intervals, serving as the foundation for both SCATE's periods and repeating intervals.

\paragraph{Period Class.} The Period class implements SCATE's period concept, encapsulating an amount of time expressed as counts of standard time units defined through the Unit class. For instance, ``three months'' is represented as \texttt{Period(MONTH, 3)}. Periods can be combined through the \texttt{PeriodSum} class to express complex durations like "two years and a day" as \texttt{PeriodSum([Period(YEAR, 2), Period(DAY, 1)])}.

\paragraph{Repeating Class.} The \texttt{Repeating} class implements SCATE's repeating interval concept to capture calendar-anchored recurring time intervals. For example, "February" (all Februaries across the timeline) is represented as \texttt{Repeating(MONTH, YEAR, value=2)}, while "Thursday" would be \texttt{Repeating(DAY, WEEK, value=3)}  (where we follow the \texttt{dateutil} library in using value=0 to represent Monday).
We also extend the Repeating class to implement common temporal concepts as specialized classes, such as \texttt{Spring}, \texttt{Summer}, \texttt{Fall}, \texttt{Winter} for seasons, and \texttt{Morning}, \texttt{Noon}, \texttt{Afternoon}, \texttt{Evening}, \texttt{Night} for parts of the day.

\subsubsection{Temporal Operators Class Definitions}

A distinctive feature of SCATE is its rich set of temporal operators. We implement these operators through a series of Python classes. 

\paragraph{Positional Operator Classes.} We implement a group of positional operators primarily used to move intervals relative to reference points. Two key operator pairs are \texttt{Last}/\texttt{Next} and \texttt{Before}/\texttt{After}. The \texttt{Last} and \texttt{Next} classes respectively find the closest intervals before or after a given Shift, while \texttt{Before} and \texttt{After} move intervals backward or forward by specific time units or occurrences. For example, "the previous summer" when spoken on February 14, 1912 can be represented as \texttt{Last(Interval.of(1912, 2, 14), Summer())}, while "three Aprils after" written on January 23, 1993 can be expressed as \texttt{After(Interval.of(1993, 1, 23), Repeating(MONTH, YEAR, value=4), n=3)}. These seemingly similar but functionally distinct operators enable us to accurately capture different types of temporal expressions in natural language.

\paragraph{Selection Operator Classes.} Selection operators primarily select specific instances from time sequences. The \texttt{Nth} class allows selection of the $n$-th shift from the beginning or end of an interval, while the \texttt{This} class finds the current shift containing a given interval. These operators apply to ordinal expressions (e.g., "the third Thursday") and deictic expressions (e.g., "this month"). For example, "third-to-last Sunday of 2024" can be represented as \texttt{Nth(Year(2024), Repeating(DAY, WEEK, value=6), index=3, from\_end=True)}, while "this January" spoken on November 10, 1037 can be expressed as \texttt{This(Interval.of(1037, 11, 10), Repeating(MONTH, YEAR, value=1))}.

\paragraph{Range Operator Classes.} Range operators handle relationships between multiple intervals. The \texttt{Between} class creates a span between two intervals, while the \texttt{Intersection} class finds the overlap among multiple intervals. These operators are particularly useful for handling time ranges and intersections of multiple temporal constraints. For example, "since 1994" written on January 9, 2007 can be represented as \texttt{Between(Year(1994), Interval.of(2007, 1, 9))}, while "earlier that day" in the context of "We met at 6:00 on January 24, 1979. Earlier that day..." would be interpreted as \texttt{Intersection([Last(Interval.of(1979, 1, 24, 6), None), Interval.of(1979, 1, 24)])}.

\paragraph{Collection Operator Classes.} We implement multiple collection operators to handle sets of intervals, such as \texttt{These}, \texttt{LastN}, and \texttt{NextN}. These operators extend the functionality of basic operators, allowing us to process multiple related intervals. For example, "the next six Fridays" when written on December 22, 1714 can be represented as \texttt{NextN(Interval.of(1714, 12, 22), Repeating(DAY, WEEK, value=4), n=6)}, while "Tuesdays in January 2025" can be expressed as \texttt{These(Interval.of(2025, 1), Repeating(DAY, WEEK, value=1))}).

\paragraph{Union and Intersection Classes.} We implement \texttt{ShiftUnion} and \texttt{RepeatingIntersection} classes that allow us to combine multiple shifts or find intersections of repeating intervals. For example, "Mondays and Fridays" can be represented as \texttt{ShiftUnion([Repeating(DAY, WEEK, value=0), Repeating(DAY, WEEK, value=4)])}, while "Saturdays in March" can be expressed as \texttt{RepeatingIntersection([Repeating(DAY, WEEK, value=5), Repeating(MONTH, YEAR, value=3)])}.

\subsection{Data Augmentation}

The expressive power and flexible design of SCATE introduces a potential challenge: the scarcity of large-scale annotated data. Under the original SCATE annotation framework, human annotators must identify all temporal expressions in text and their corresponding operators, as illustrated in \Cref{fig:scate-example}. This precise annotation requires domain experts; otherwise, annotation quality may be insufficient for model training. For instance, \citet{su-etal-2021-university} report that even when two PhD students from related fields spent approximately 10 days on annotations, those annotations ultimately degraded model performance in temporal expression recognition due to lack of annotator training on the complex SCATE annotation guidelines.

Given our complete Python implementation of SCATE, a natural question arises: since LLMs like Claude 3.7 \citep{claude37sonnet} and GPT-4.1 \citep{openai2025gpt41} have demonstrated unprecedented code generation capabilities, could we leverage them to identify temporal expressions in unlabeled text and generate corresponding SCATE Python code at scale? This would yield (text, time expression, python code) triplets for training smaller, more deployable language models.

To this end, we construct detailed LLM prompts (SCATE prompt) for data augmentation. We present our data augmentation prompts in markdown format, with an example shown in the \Cref{sec:scate-prompt}. These prompts effectively serve as formal documentation for our Python library, thoroughly introducing SCATE's key temporal concepts, defining each implemented class, detailing possible usage patterns, and providing examples to clarify potential confusion points (e.g., distinguishing between Next and After operators). Our objective is to fully leverage LLMs' code generation capabilities in a code-generation framework, adapting them to generate code for Python libraries they may not have encountered during pre-training.

An immediate post-generation constraint enforcement method emerges from our use of well-defined Python code objects as targets: we simply execute the generated SCATE Python code and discard samples that produce runtime errors, thus ensuring syntactically and semantically valid SCATE Python code.

\section{Experiments}
\label{sec:experiments}
\subsection{Datasets}
TempEval-2013 \citep{uzzaman-etal-2013-semeval} data has been annotated with publicly available SCATE annotations, including training, development, and test sets. Using our implemented SCATE Python library, we convert the original XML-formatted SCATE annotations into (sentence, time expression, SCATE code) triplets. The objective is to identify all potential time expressions in input sentences and represent them using SCATE code. To obtain a larger test set for evaluating the generalization capability of our proposed method, we merge the original development and test sets into a consolidated test set for final model performance evaluation. The resulting dataset includes 557 annotated SCATE code block in the training set and 313 in the test set. We use the training set to refine prompts and select the optimal LLM for data augmentation.

\paragraph{Evaluation Metrics}
For evaluation, we assess each gold-standard triplet's time expression by first determining whether it was successfully identified. If a time expression remains undetected, its normalization accuracy is assigned a value of 0. For identified expressions, we execute the predicted SCATE code to generate normalized time intervals and compare them with the gold-standard intervals. We assign an accuracy of 1 for exact matches and 0 otherwise. Based on this approach, we calculate standard evaluation metrics including average execution accuracy, precision, recall, and F1 score.

\subsection{Implementation Details}
\paragraph{Models and Hyperparameters} We access various LLMs through cloud-based APIs: GPT-4.1 via Azure OpenAI, Claude 3.7 via Amazon Bedrock, and Gemini \citep{gemini_flash, deepmind2025geminipro} via Google Cloud Platform. For local model training, we utilize Qwen/Qwen2.5-0.5B-Instruct, which we train on a single NVIDIA 80GB A100 GPU. Our training hyperparameters include a learning rate of $2 \times 10^{-5}$, 5 training epochs, and a batch size of 64. We prompt or train models to generate JSON strings where each item represents a time expression and SCATE code pair, for example: \texttt{[{"\{time\_text": "recent years", "scate": "Last(interval=Interval.of(1998, 2, 13), shift=Period(unit=YEAR, n=None))\}"}]}.

\paragraph{Data Augmentation Text}
We randomly sample 10k sentences from the CC-News \citep{mackenzie2020cc} dataset—widely used in large language model pretraining—as our source for data augmentation. It comes from the same newswire domain as with TempEval SCATE-annotated data. We then apply Claude 3.7 to these sentences using our designed SCATE prompt to generate temporal annotations. After prompt-based generation, we apply a runtime filtering step to discard syntactically invalid or semantically incoherent outputs. This process yields a total of 8,583 valid SCATE code blocks.

\subsection{Main Results}

\begin{table}[htbp]
   \centering
   \caption{Performance of LLMs on Temporal Expression Recognition and SCATE Code Generation on Training Set.}
   \label{tab:llm_performance}
   \begin{tabular}{lcccc}
       \toprule
       \textbf{Model} & \textbf{Accuracy} & \textbf{Precision} & \textbf{Recall} & \textbf{F1} \\
       \midrule
       Claude 3.5     & 0.62            & \textbf{0.64} & 0.62          & 0.63 \\
       Claude 3.7     & \textbf{0.69}   & 0.63          & 0.69          & \textbf{0.66} \\
       Gemini 2.0 Flash & 0.64          & \textbf{0.64} & 0.64          & 0.64 \\
       Gemini 2.5 Flash & 0.61          & 0.63          & 0.61          & 0.62 \\
       Gemini 2.5 Pro & 0.50            & \textbf{0.64} & 0.50          & 0.56 \\
       GPT-4.1        & 0.67            & 0.60          & \textbf{0.67} & 0.63 \\
       \midrule
       Average        & 0.62 & 0.63 & 0.62 & 0.62 \\
       \bottomrule
   \end{tabular}
\end{table}

\begin{table}[htbp]
  \centering
  \caption{Performance comparison of different methods on the test set.}
  \label{tab:test-set-performance}
  \begin{tabular}{lcccc}
    \toprule
    \textbf{Methods}  & \textbf{Accuracy} & \textbf{Precision} & \textbf{Recall} & \textbf{F1} \\
    \midrule
    Qwen2.5-0.5B + Training Set & 0.01 & 1.00 & 0.01 & 0.01 \\
    Qwen2.5-0.5B + CC-News & 0.37 & 0.46 & 0.37 & 0.41 \\
    Qwen2.5-0.5B + CC-News + Training Set & \textbf{0.59} & \textbf{0.59} & \textbf{0.59} & \textbf{0.59} \\
    Claude 3.7 + SCATE Prompt & 0.49 & 0.56 & 0.49 & 0.52 \\
    Claude 3.7 + Interval Few-shot Prompt & 0.38	& 0.38 & 0.38 & 0.38 \\
    GPT 4.1 + SCATE Prompt & 0.51 & 0.51 & 0.51 & 0.51 \\
    \bottomrule
  \end{tabular}
  
\end{table}

\paragraph{Can LLMs identify time expressions and parse corresponding SCATE code?}
To answer this question and identify the optimal state-of-the-art model for our designed data augmentation, we test the most popular large language models on the training set, including Claude 3.5 Haiku, Claude 3.7 Sonnet, Gemini 2.0 Flash, Gemini 2.5 Flash, Gemini 2.5 Pro, and GPT 4.1. We prompt these models using the approach described in Section 3.3 to identify temporal expressions and generate corresponding SCATE code. The results are presented in \Cref{tab:llm_performance}. These state-of-the-art models achieve an average accuracy and F1 score of 0.62, indicating that while large models can generate SCATE code through prompting to a reasonable degree, there remains significant room for improvement. Among the evaluated models, Claude 3.7 demonstrates the best performance.

\paragraph{Is targeting SCATE code better than directly generating time intervals?}

One of our key hypotheses for time normalization is that existing LLMs can more effectively generate code representations of temporal operations than implicitly perform these operations to directly identify time expressions and their corresponding intervals. Similar to recent work \citep{wei2022chain} finding that generating chain-of-thought reasoning before providing answers significantly improves performance in mathematical reasoning, we posit that SCATE code acts as a form of chain-of-thought, with the additional benefit that we need not rely on the model to produce the final answer since we can determine this through code execution.

To validate this hypothesis, we employ conventional few-shot prompting (Interval Few-shot Prompt) to direct Claude 3.7 in identifying temporal expressions and their corresponding time intervals in input text (the specific Interval Few-shot Prompt is provided in \Cref{{sec:few-shot-prompt}}). We contrast this with prompting Claude 3.7 to generate SCATE code using our approach described in Section 3.3. As shown in \Cref{tab:test-set-performance}, the SCATE code generation approach significantly outperforms direct time interval generation, surpassing it by more than 10 points in both average accuracy and F1 scores.

\paragraph{Can we train a smaller local model for SCATE generation?}

Temporal normalization typically functions as one component within larger data processing or retrieval pipelines. Consequently, there is a clear need for efficient solutions with low computational costs. Relying entirely on large models like Claude 3.7 presents challenges for scaling temporal normalization deployments. Therefore, a smaller model ($\leq$ 1B parameters) capable of inference on a single consumer-grade GPU better aligns with practical requirements.

We explore this possibility by fine-tuning Qwen2.5-0.5B on three different data combinations: on the original TempEval 2013 training set (Training Set), on data augmentation-labeled data (CC-News), and on a combination of both (CC-News + Training Set). We observe that small models struggle to achieve reasonable performance on limited datasets, even with manually annotated training data from TempEval, with average accuracy and F1 scores approaching zero. Fine-tuning on augmentation-labeled data shows notable improvement, though still maintaining a considerable gap compared to its parent data augmentation method (Claude 3.7 + SCATE Prompt).

However, when we combine our augmented data with the original training set, we observe significant performance improvements, achieving 0.59 average accuracy on the test set—surpassing its parent data augmentation method (Claude 3.7 + SCATE Prompt) by 10 points. This demonstrates the complementary nature of augmented and original training data, and confirms that smaller, deployable models can effectively perform temporal normalization when provided with sufficient and diverse training examples.

\paragraph{What are the main errors in the best-performing fine-tuned model?}

We conducted an error analysis on 20 errors from the Qwen2.5-0.5B + CC-News + Training Set model. The analysis reveals that the most common issue is missed temporal expressions (70\%), where the system fails to identify time phrases annotated in the gold annotations. The second most frequent problem is boundary errors (10\%), where the system's identified temporal expression has boundary differences from the gold standard, despite the SCATE expression being logically similar. For example, the gold standard might annotate ``year,'' while the system identifies ``within a year.''

Some errors (10\%) are structural, such as in ``third-quarter net loss... year-earlier,'' which involves nested temporal comparisons (using Before(Nth(...)) structure), but the system only captures individual fragments rather than the complete structure. SCATE type errors (5\%) manifest as inappropriate operator selection, such as labeling ``later'' as Next(...) instead of the more semantically appropriate After(...). Granularity errors (5\%) occur when the system represents specific expressions with coarser time units, such as simplifying ``11/02/89'' to Year(1989) instead of preserving the month and day information.

Overall, the primary challenges lie not in the model's ability to generate SCATE code but in span recognition—a traditional NLP task. Improving the model's identification of temporal expression spans represents a promising research direction for enhancing overall performance, or alternatively, introducing an additional small classification model specifically for span identification.

\section{Limitations}
\label{sec:limitations}
We focus on TempEval-2013, the only publicly available dataset with full SCATE annotations, as the primary benchmark for evaluation. While this dataset provides high-quality supervision, it does not fully capture the diversity of temporal expressions found in open-domain or multilingual scenarios. Our model fine-tuning experiments are limited to a single small-scale open-source model (Qwen2.5-0.5B), without exploring alternative architectures or larger models. Additionally, our data augmentation pipeline relies on a subset of proprietary LLMs, and we do not systematically compare across the full range of commercial or open-source models.

\section{Conclusion}
We present a new end-to-end approach for time normalization by framing it as a code generation problem based on the SCATE framework. Our method unifies symbolic temporal semantics with executable representations, enabling deterministic and interpretable normalization of complex time expressions. Through comprehensive Python implementation and carefully designed prompting strategies, we show that LLMs can effectively identify temporal expressions and generate high-quality SCATE code. More importantly, we demonstrate that fine-tuning small models on a combination of LLM-augmented and human-annotated data achieves strong performance while remaining deployable on standard hardware. Our findings suggest that compositional code generation offers a scalable and semantically grounded solution for time normalization.

\bibliographystyle{plainnat}
\bibliography{anthology,custom}







\appendix
\newpage

\clearpage
\setcounter{section}{0}
\setcounter{equation}{0}
\setcounter{figure}{0}
\setcounter{table}{0}

\renewcommand{\thesection}{\Alph{section}}
\renewcommand\thefigure{S\arabic{figure}}
\renewcommand\thetable{S\arabic{table}}
\renewcommand\theequation{S\arabic{equation}}


\section{AI Use Declaration} 
\label{sec:ai_use}
Code was developed with support from GitHub Copilot. ChatGPT was used for editing for grammar and clarity in some sections.

\section{SCATE Prompt}
\label{sec:scate-prompt}
\includepdf[pages=-,scale=0.85,pagecommand={}]{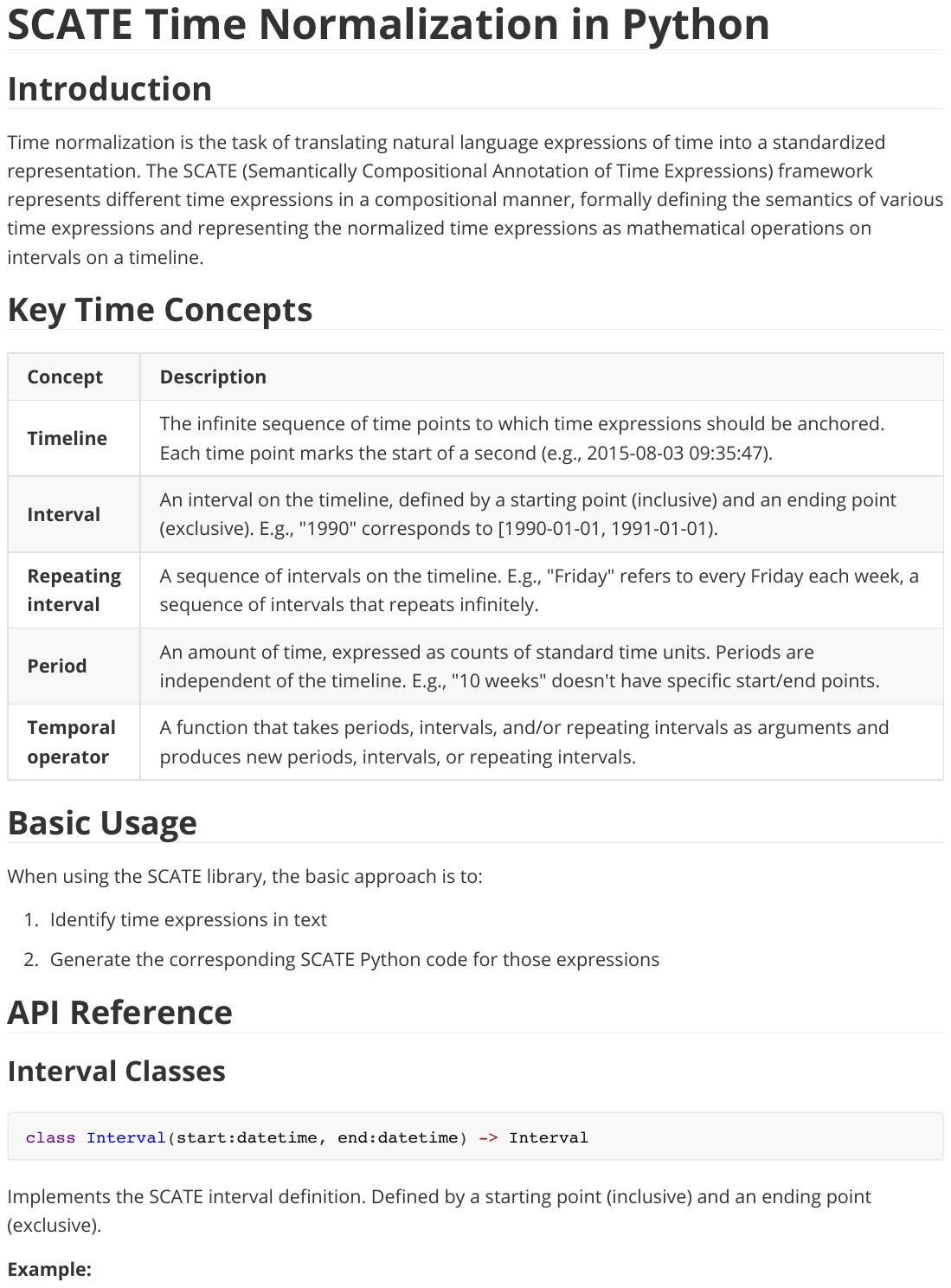}

\section{Interval Few-shot Prompt}
\label{sec:few-shot-prompt}
\includepdf[pages=-,scale=0.85,pagecommand={}]{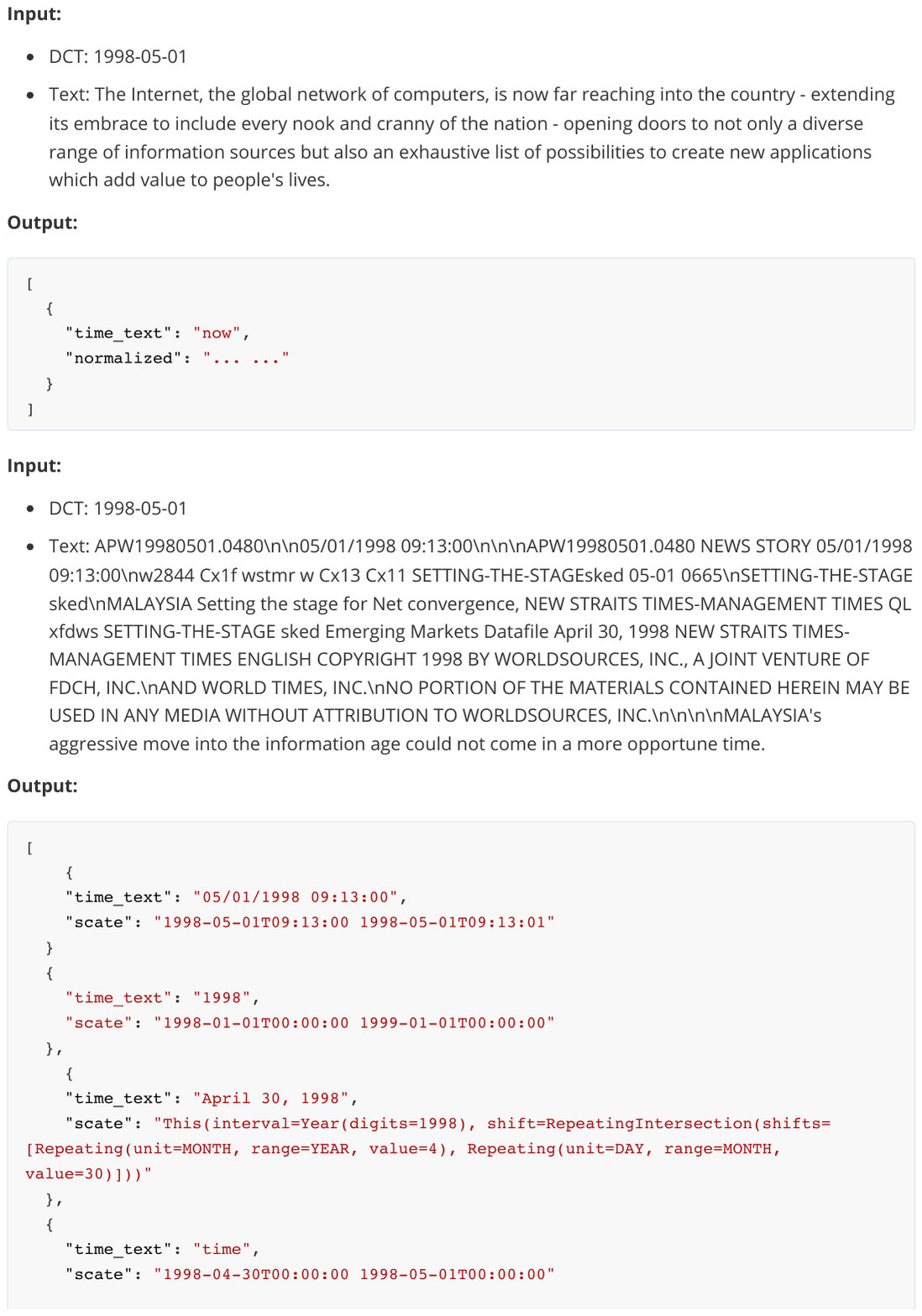}

\section{Confidence Intervals}

We calculate 95\% confidence intervals using bootstrap sampling, where we randomly resample 80\% of the evaluation data 100 times and compute evaluation metrics on each sample to estimate the statistical uncertainty of our model's performance. We present the results in \Cref{tab:llm_performance_ci} and \Cref{tab:test-set-performance-ci}.

\begin{table}[htbp]
   \centering
   \caption{Performance of LLMs on Temporal Expression Recognition and SCATE Code Generation on Training Set with 95\% confidence intervals. Values shown as mean ± std, confidence intervals (CI): (lower, upper).}
   \label{tab:llm_performance_ci}
   \resizebox{\textwidth}{!}{%
   \begin{tabular}{lcccc}
       \toprule
       \textbf{Model} & \textbf{Accuracy} & \textbf{Precision} & \textbf{Recall} & \textbf{F1} \\
       \midrule
       Claude 3.5 & 0.62 ± 0.02 & 0.64 ± 0.02 & 0.62 ± 0.02 & 0.63 ± 0.02 \\
       & CI: (0.62, 0.63) & CI: (0.64, 0.65) & CI: (0.62, 0.63) & CI: (0.63, 0.64) \\
       \midrule
       Claude 3.7 & \textbf{0.69 ± 0.02} & 0.63 ± 0.02 & \textbf{0.69 ± 0.02} & \textbf{0.66 ± 0.02} \\
       & \textbf{CI: (0.69, 0.70)} & CI: (0.63, 0.64) & \textbf{CI: (0.69, 0.70)} & \textbf{CI: (0.66, 0.67)} \\
       \midrule
       Gemini 2.0 Flash & 0.65 ± 0.03 & 0.64 ± 0.02 & 0.65 ± 0.03 & 0.64 ± 0.03 \\
       & CI: (0.64, 0.65) & CI: (0.63, 0.64) & CI: (0.64, 0.65) & CI: (0.64, 0.65) \\
       \midrule
       Gemini 2.5 Flash & 0.61 ± 0.02 & 0.63 ± 0.02 & 0.61 ± 0.02 & 0.62 ± 0.02 \\
       & CI: (0.61, 0.62) & CI: (0.63, 0.64) & CI: (0.61, 0.62) & CI: (0.62, 0.63) \\
       \midrule
       Gemini 2.5 Pro & 0.50 ± 0.02 & \textbf{0.65 ± 0.02} & 0.50 ± 0.02 & 0.56 ± 0.02 \\
       & CI: (0.50, 0.51) & \textbf{CI: (0.64, 0.65)} & CI: (0.50, 0.51) & CI: (0.56, 0.57) \\
       \midrule
       GPT-4.1 & 0.66 ± 0.02 & 0.60 ± 0.02 & 0.66 ± 0.02 & 0.63 ± 0.02 \\
       & CI: (0.66, 0.67) & CI: (0.60, 0.60) & CI: (0.66, 0.67) & CI: (0.63, 0.63) \\
       \bottomrule
   \end{tabular}
     }
\end{table}

\begin{table}[htbp]
  \centering
  \caption{Performance comparison of different methods on the test set with 95\% confidence intervals. Values shown as mean ± std, confidence intervals (CI): (lower, upper).}
  \label{tab:test-set-performance-ci}
  \resizebox{\textwidth}{!}{%
  \begin{tabular}{lcccc}
    \toprule
    \textbf{Methods}  & \textbf{Accuracy} & \textbf{Precision} & \textbf{Recall} & \textbf{F1} \\
    \midrule
    Qwen2.5-0.5B + Training Set & 0.01 ± 0.01 & 0.49 ± 0.50 & 0.01 ± 0.01 & 0.01 ± 0.01 \\
    & CI: (0.00, 0.01) & CI: (0.39, 0.59) & CI: (0.00, 0.01) & CI: (0.01, 0.01) \\
    \midrule
    Qwen2.5-0.5B + CC-News & 0.37 ± 0.03 & 0.46 ± 0.04 & 0.37 ± 0.03 & 0.41 ± 0.03 \\
    & CI: (0.36, 0.37) & CI: (0.45, 0.46) & CI: (0.36, 0.37) & CI: (0.40, 0.41) \\
    \midrule
    Qwen2.5-0.5B + CC-News + Training Set & \textbf{0.59 ± 0.04} & \textbf{0.59 ± 0.04} & \textbf{0.59 ± 0.04} & \textbf{0.59 ± 0.04} \\
    & \textbf{CI: (0.58, 0.60)} & \textbf{CI: (0.59, 0.60)} & \textbf{CI: (0.58, 0.60)} & \textbf{CI: (0.58, 0.60)} \\
    \midrule
    Claude 3.7 + SCATE Prompt & 0.49 ± 0.03 & 0.56 ± 0.04 & 0.49 ± 0.03 & 0.52 ± 0.03 \\
    & CI: (0.49, 0.50) & CI: (0.55, 0.57) & CI: (0.49, 0.50) & CI: (0.52, 0.53) \\
    \midrule
    Claude 3.7 + Interval Few-shot Prompt & 0.38 ± 0.03 & 0.39 ± 0.03 & 0.38 ± 0.03 & 0.38 ± 0.03 \\
    & CI: (0.38, 0.39) & CI: (0.38, 0.39) & CI: (0.38, 0.39) & CI: (0.38, 0.39) \\
    \midrule
    GPT 4.1 + SCATE Prompt & 0.51 ± 0.03 & 0.51 ± 0.03 & 0.51 ± 0.03 & 0.51 ± 0.03 \\
    & CI: (0.51, 0.52) & CI: (0.50, 0.52) & CI: (0.51, 0.52) & CI: (0.50, 0.52) \\
    \bottomrule
  \end{tabular}%
  }
\end{table}

\section{License information}
We respect the license and intended use of all models and datasets employed in this study. Detailed license information is provided below.

\paragraph{Models.}
The Claude family models utilized in our study are licensed under the \href{https://www.anthropic.com/legal/commercial-terms}{Commercial Terms of Service}. The Gemini family models are licensed under the \href{https://developers.google.com/terms}{Google APIs Terms of Service}. The GPT-4.1 model is licensed under the \href{https://openai.com/policies/business-terms/?utm_source=chatgpt.com}{Business terms}. The Qwen 2.5 models are licensed under the \href{https://choosealicense.com/licenses/apache-2.0/}{Apache License 2.0}.

\paragraph{Datasets.}
The CC-News dataset used in our study is available under the \href{https://commoncrawl.org/terms-of-use}{Common Crawl Terms of Use}.

\end{document}